\documentclass[12pt]{article}
\usepackage[utf8]{inputenc}
\usepackage{amsmath,amssymb,amsfonts}
\usepackage{graphicx}
\usepackage{booktabs}
\usepackage{caption}
\usepackage{hyperref}
\usepackage{xcolor}
\usepackage{multirow}
\usepackage{array}
\usepackage{float}

\usepackage[backend=biber, 
            sorting=none,
            maxnames=99]{biblatex}
\nocite{*}

\title{Attention's Gravitational Field: \newline
\fontsize{14}{16}\selectfont  A Power-Law Interpretation of Positional Correlation }
\author{Edward Zhang \footnote{contact with:windyrobin@aliyun.com}}
\date{}
\begin{document}

\maketitle

\begin{abstract}
This paper explores the underlying principles of positional relationships and encodings within Large Language Models (LLMs) and introduces the concept of Attention-Gravitational Field (AGF). By decoupling positional encodings from semantic embeddings, we optimize the model architecture and achieve superior accuracy compared to prevailing encoding methods. Furthermore, we provide an in-depth analysis of AGF, demonstrating its intrinsic consistency with learning and stability curves, as well as its empirical alignment with Newton’s Law of Universal Gravitation. By offering a rigorous theoretical exploration of these phenomena, this work represents a significant step toward interpreting the Attention mechanism and unlocks new possibilities for future research in model optimization and interpretability.
\end{abstract}

\section{Introduction}
Early LLM models predominantly adopted absolute positional encodings of Transformer\cite{vaswani2017attention}, which were fused directly with semantic embeddings. This approach is conceptually perplexing—much like summing `age' and `income' into a single value, which inevitably leads to semantic distortion and confusion. Despite this, it remains the prevailing method today due to its robust empirical performance and high computational efficiency.

\vspace{0.5em}
In recent years, alternative explorations have emerged. RoPE \cite{su2024roformer}, for instance, attempts to confine this semantic fusion to specific stages. Other methods, such as T5\cite{raffel2020exploring}, employ relative positional encoding but still require significant parameter overhead. Furthermore, driven by the demand for model extrapolation, works like ALiBi\cite{press2021train} have simplified relative positioning to a parameter-free form, while KERPLE \cite{chi2022kerple}provides a suite of formulations through kernel-based approaches.

\vspace{0.5em}

While these contributions are highly insightful and have informed our research, two major issues remain: firstly, their comprehensive performance in production environments often falls short of absolute positional encoding; secondly, and most importantly, they lack an explanation for the ``Why". They fail to address the most fundamental question: What is the underlying essence of positional relationships?

\section{Methodology and Background}

We focus on optimizing the internal logic of the standard Attention calculation:
$$a_{m,n}=\frac{\exp(q_{m}^{\top}k_{n}/\sqrt{d})}{\sum_{i=1}^{L}\exp(q_{m}^{\top}k_{i}/\sqrt{d})}, \quad o_{m}=\sum_{n=1}^{L}a_{m,n}v_{n}$$

\vspace{1em}

\begin{table}[htbp]
\centering
\caption{Summary of Experimental Configurations and Hardware Environment.}
\label{tab:experimental_setup}
\begin{tabular}{ll}
\hline
\textbf{Configuration Item} & \textbf{Value / Description} \\ \hline
Base Architecture           & Vanilla Transformer (Transformer-BIG) \\
Dataset                     & WMT\_17 (Translation Task, en-de) \\
Number of Layers            & 3 (Reduced from default 6) \\
Precision                   & FP16 (Half-Precision) Training \\
Hardware                    & Single NVIDIA Tesla V100-PCIE-32GB \\
Training Duration           & $\approx$ 15 hours per run \\
\hline
\end{tabular}
\end{table}

To accelerate the training process, we employed FP16(Half-Precision)training, the number of layers(LayerNum) was set to 3, rather than the default 6.

\vspace{1em}
Training framework  \footnote{ OpenNMT-py: \url{https://github.com/OpenNMT/OpenNMT-py}}  and  AGF module code  \footnote{AGF git: \url{https://github.com/windyrobin/AGF/tree/main}} is avaiable now.

\section{Positional Correlation}
\subsection{Decomposition}
According to the Attention formula, when two tokens (or embeddings) at different positions calculate their $QK$ relationship, we posit that the role of positional encoding consists of two distinct components based on heuristic intuition:

\begin{enumerate}
    \item \textbf{Relative Positional Component:} This depends exclusively on the relative distance between the two tokens.
    \item \textbf{Absolute Positional Component:} This accounts for specific indices (e.g., the first or last token in a sentence) that might carry unique significance.
\end{enumerate}

However, we argue that the \textit{Relative Component} (\#1) captures almost all valuable information, while the \textit{Absolute Component} (\#2) is virtually negligible. Although various encoding methods—such as \textbf{ALiBi, T5, RoPE,} and \textbf{Sinusoidal}(Sinu)—employ different techniques, their essence is to ensure that for every feature dimension (e.g., $d=1024$) and every relative position (e.g., $SeqLen=128$) within an Attention layer, there are specific parameters or coefficients. 

\vspace{.5em}

In this context, the total parameter space can be expressed as:
\begin{equation}
\text{Total Parameters} = H \times D_k \times \text{SeqLen} \times 2
\end{equation}

Therefore, when employing a relative positional encoding scheme, the Attention computation allows for the decoupling of the relative positional component from the intrinsic token features:

\begin{equation}
a_{m, n}=\frac{\exp\left(q_{m}^{\top} k_{n}/\sqrt{d} \times \text{PosCoeff}\right)}{\sum_{i=1}^{L}\exp\left(q_{m}^{\top} k_{i}/\sqrt{d} \times \text{PosCoeff}\right)},\quad o_{m}=\sum_{n=1}^{L} a_{m, n}v_{n}.
\end{equation}

Notably, in contrast to existing methods like \textbf{ALiBi} and \textbf{T5}, we adopt a \textbf{multiplicative} interaction rather than an additive bias. While addition in the logit space is mathematically related to multiplication after the $\exp$ operation, the distinction lies in the stage at which positional information is integrated. We contend that our multiplicative approach is more theoretically sound, a claim for which we provide a rigorous justification in the subsequent sections.

\subsection{Directionality}
In the aforementioned formula, the parameter count is multiplied by a factor of 2 to account for the \textbf{directionality} of relative positions. Specifically, a relative distance of $+3$ versus $-3$ should correspond to distinct coefficients. 

This is intuitively grounded in linguistic structures. Consider the relationship between the words \textit{`beautiful'} and \textit{`girl'}:
\begin{itemize}
    \item As a pre-modifier (e.g., \textit{``there is a beautiful girl"}), the relative distance might be $1$.
    \item As a post-modifier (e.g., \textit{``the girl is beautiful"}), the words are typically separated by a linking verb (e.g., \textit{`is'}), increasing the relative distance.
\end{itemize}
These two patterns---"pre-modifier" vs. "post-modifier"---represent different syntactic structures that may be captured by distinct Attention heads. Consequently, it is essential for relative position computations to be \textbf{direction-aware}. 

\vspace{.5em}

While Causal Language Models (CLMs) bypass this requirement because tokens cannot attend to subsequent positions, it remains critical for Translation Models (Encoder-Decoder or Bi-directional architectures). Our empirical results confirm that incorporating bi-directional directionality yields significant performance gains.

\subsection{Component Analysis (LC 1-3)}
Each attention head, typically comprising 64 dimensions, represents a cluster of correlated features (justifying the use of additive operations within the Attention mechanism). Inspired by the principles of \textbf{Principal Component Analysis (PCA)} and \textbf{Fourier Decomposition}, we decompose the positional influence into three hierarchical components. These range from low to high frequency, or from coarse to fine-grained granularity, with a corresponding increase in parameter volume and information capacity:

\begin{itemize}
    \item \textbf{LC\_1 (Layer Component 1):} Each head ($H$) is treated as a holistic unit with a single parameter, characterizing a macroscopic decay curve relative to distance.
    \begin{itemize}
        \item Parameter Count: $2 \times H$ (accounting for bidirectional symmetry).
    \end{itemize}
    
    \item \textbf{LC\_2 (Layer Component 2):} Within each head, an amplitude parameter (centered at 1) is assigned to each relative position ($pos < \text{SeqLen}$).
    \begin{itemize}
        \item Parameter Count: $2 \times H \times \text{SeqLen}$.
    \end{itemize}
    
    \item \textbf{LC\_3 (Layer Component 3):} A fine-grained weight parameter (centered at 1) is assigned to every individual feature dimension within each head for every relative position.
    \begin{itemize}
        \item Parameter Count: $2 \times H \times d_k \times \text{SeqLen}$.
    \end{itemize}
\end{itemize}

\begin{figure}
    \centering
    \includegraphics[width=0.8\linewidth]{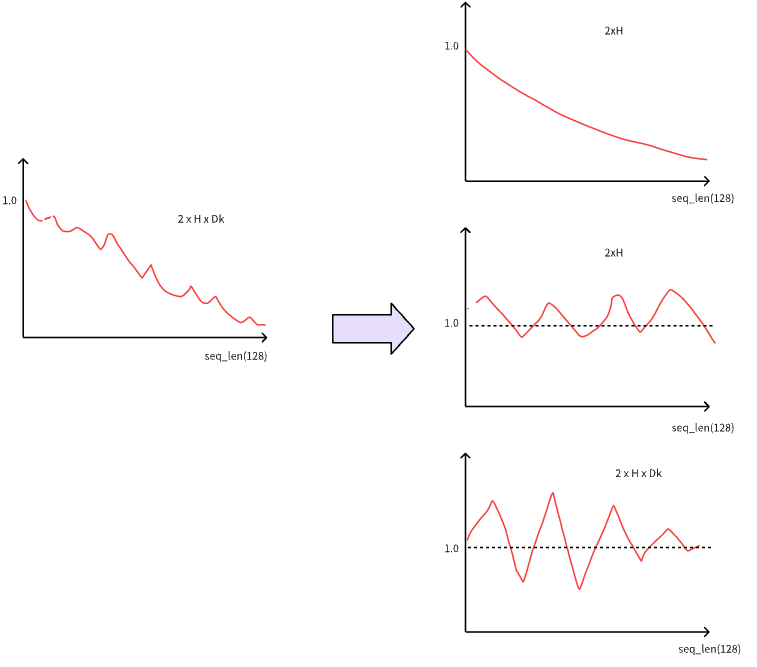}
    \caption{Decomposition}
    \label{fig:placeholder}
\end{figure}
The final positional coefficient is defined as the product of these hierarchical components:
\begin{equation}
    \text{PositionCoeff}(A_m, B_n) = LC_1(A_m, B_n) \cdot LC_2(A_m, B_n) \cdot LC_3(A_m, B_n)
\end{equation}

\textit{Note: Components $LC_1$ and $LC_2$ can be integrated into a unified $LC_2$ representation during training. However, if $LC_2$ demonstrates negligible contribution to the model's performance, the architecture can be further simplified by retaining only $LC_1$.}

\section{Attention's Gravitational Field}

For the fitting of the \textbf{LC-1} component, we seek a non-linear decay curve relative to distance. Intuitively, the interaction strength---whether interpreted as affinity or attention---between two tokens should exhibit non-linear attenuation. This naturally evokes an analogy with \textbf{Newton’s Law of Universal Gravitation}:

\begin{equation}
    F = G \frac{M \cdot m}{r^2}
\end{equation}

By treating the $Q$ and $K$ vectors as masses $M$ and $m$, and considering an object at altitude $d$ above a sphere (e.g., Earth) with radius $r$, the force is given by:
\begin{equation}
    F(M, m, d) = \frac{G \cdot M \cdot m}{(r + d)^2}
\end{equation}

In translation tasks (e.g., WMT) where the relative distance $d$ resides in $[0, 128]$, let $Base = F(0) = \frac{G \cdot M \cdot m}{r^2}$. The relative decay can then be formulated as:
\begin{equation}
    F(d) = Base \cdot \frac{1}{(1 + d/r)^2}
\end{equation}

To provide a generalized form for multi-dimensional spaces, we define:
\begin{equation}
    F(d) = Base \cdot \frac{1}{(1 + d/r)^k}, \quad k > 0, r > 0, d \geq 0
\end{equation}

Heuristic observation suggests that typical sentence lengths range from 8 to 40 tokens, with basic syntactic dependencies involving a minimum of 2 tokens. To ensure a robust convergence baseline, we initialize the reciprocal radius at $(1/r) = 1/24$, providing a non-restrictive initial state for the field. To maintain direction-awareness, each attention head per layer utilizes two trainable parameters ($G$ and $r$) for each direction, resulting in $4 \times H$ parameters per layer.
\newline
We term this approach \textbf{AGF (Attention-Gravitational Field)}, corresponding to our \textbf{LC-1} granularity. When AGF is used as a prior to jointly fit the $LC_1$ and $LC_2$ levels, the method is denoted as \textbf{AGF-M (Middle)}. To validate these methods, we removed the absolute Sinusoidal Positional Encoding (Sinu PE) and compared the performance of \textbf{AGF}, \textbf{AGF-M}, and the \textbf{Vanilla Transformer}. The results are summarized below:
\begin{table}[h]
\centering
\caption{Performance Comparison of Different Positional Encoding Methods}
\begin{tabular}{cccc}
\toprule
Mode & Use PE & Position Mode & Validation Accuracy \\
\midrule
\multirow{3}{*}{QKV} & Yes & --& 70.5911 \\
\cline{2-4}
 & \multirow{2}{*}{No} & AGF & 70.4511 \\
 &  & AGF-M & 70.4863 \\
\bottomrule
\end{tabular}
\end{table}

As illustrated in the results, the performance difference between \textbf{AGF} and \textbf{AGF-M} is nearly negligible. However, there is a marginal decrease compared to the \textbf{Vanilla} baseline, approximately $-0.15$ on a scale of $70$.

\vspace{.5em}

Intriguingly, following the formalization and testing of AGF, we encountered the \textbf{KERPLE}\cite{chi2022kerple} framework through a post-hoc literature review. KERPLE rigorously compares T5, ALiBi, and RoPE, proposing a high-performing composite kernel defined as:
\begin{equation}
    k^{\mathrm{comp}}([\boldsymbol{q}_{m}, m], [\boldsymbol{k}_{n}, n]) = \boldsymbol{q}_{m}^{\top} \boldsymbol{k}_{n} + c - r_{1} \cdot \log(1 + r_{2} |m - n|), \quad r_{1}, r_{2} > 0
\end{equation}

\vspace{.5em}

By exponentiating the positional bias term, one can observe a striking mathematical convergence: $e^c$ corresponds to our $G$ parameter, $r_1$ maps to our spatial dimension $k$, and $r_2$ is analogous to the reciprocal of our radius $1/r$. This confirms that the intrinsic logic of AGF is fundamentally aligned with the kernelized approach.

\vspace{.5em}
The extensive benchmarks provided by KERPLE\cite{chi2022kerple} across datasets such as \textbf{OpenWebText2, GitHub, and ArXiv} provide strong corroborating evidence for the efficacy and generalizability of the AGF mechanism. Unlike the complex kernel-engineering process described in KERPLE, our AGF derivation is notably \textbf{simple and elegant}. Most significantly, its alignment with the \textbf{Newtonian physical laws} invites a deeper inquiry into the underlying nature of Attention mechanisms.

\vspace{.5em}
While the current results are encouraging, they prompt a pivotal question: is there further potential for optimization?

\section{Benefits of Decoupling}
\subsection{PCM-V: Positional Coefficient Multiplication of $V$}

In conventional PE frameworks, the entanglement of positional and semantic information poses significant challenges to \textbf{interpretability} and hinders semantic-level optimizations. By utilizing \textbf{AGF/AGF-M}, we decouple these components, thereby enabling novel architectural refinements. Below, we describe a key optimization that substantially enhances model accuracy.

We re-evaluate the fundamental Attention mechanism:
\begin{equation}
    a_{m,n}=\frac{\exp \left( \boldsymbol{q}_{m}^{\top} \boldsymbol{k}_{n} / \sqrt{d} \cdot \text{PosCoeff}\right)}{\sum_{i=1}^{L} \exp \left( \boldsymbol{q}_{m}^{\top} \boldsymbol{k}_{i} / \sqrt{d} \cdot \text{PosCoeff}\right)}, \quad \boldsymbol{o}_{m}=\sum_{n=1}^{L} a_{m, n} \boldsymbol{v}_{n}
\end{equation}

The relative positional coefficient essentially represents a scaling effect on the projection of Key and Value vectors relative to the Query's position. While current methods apply this scaling to the Attention weights, they typically neglect its influence during the final Value aggregation stage. We argue that for theoretical consistency, the final output must also incorporate these positional constraints.

\vspace{.5em}
This is easy to understand. Let’s assume there are two Values/Tokens, and their raw values, positional coefficients, projection scores, and normalized scores are as follows:
\begin{table}[h]
\centering
\caption{Output Contribution Calculation}
\begin{tabular}{ l  l  l  l  l }
\hline
\textbf{Raw-K/V} & \textbf{PosCoeff} & \textbf{raw\_score} & \textbf{softmax\_score} & \textbf{Output}  \\
\hline
100 & 0.1 & 100*0.1=10 & 0.1 & 0.1 * 100 ? \\
\hline
20 & 0.5 & 20*0.5=10 & 0.1 & 0.1 * 20 ? \\
\hline
\end{tabular}
\end{table}
\newline
Therefore, when calculating the contribution to the output, simply multiplying by the Raw-V is clearly unreasonable; yet, this is a pervasive issue present in all current LLM models.

\vspace{.5em}
Consequently, we propose a revised aggregation step:
\begin{equation}
    \boldsymbol{o}_{m}=\sum_{n=1}^{L} a_{m, n} \cdot \text{PosCoeff} \cdot \boldsymbol{v}_{n}
\end{equation}

Empirical tests demonstrate that this modification leads to a marked improvement in accuracy, yielding gains of \textbf{0.25--0.35} (relative to a 70-point baseline). This improvement enables our approach to outperform the \textbf{Vanilla Absolute Positional Encoding} method. We refer to this optimized variant as \textbf{PCM-V (Positional Coefficient Multiplication of Value)}.

\begin{table}[h]
\centering
\caption{PCM-v Optimization Effect}
\begin{tabular}{c|ccc}
\toprule
\textbf{Mode} & \textbf{Pos-Correlation} & \textbf{PCM-V} & \textbf{Validation Accuracy (\%)} \\
\midrule
QKV-Vanilla/PE & -- & --  & 70.5911 \\
\hline
\multirow{4}{*}{QKV} & \multirow{2}{*}{PGH} & -- & 70.4511 \\
 &  & Yes & 70.7305 \\
 \cline{2-4}
 & \multirow{2}{*}{PGH-M} & -- & 70.4863 \\
 &  & Yes & 70.7582 \\
\bottomrule
\end{tabular}
\end{table}

\vspace{.5em}
To further validate our hypothesis, we conducted comparative benchmarks using \textbf{ALiBi}. We implemented a bidirectional and learnable variant, designated as \textbf{ALiBi-B-L} (conceptually analogous to the \textit{Power} kernel in KERPLE). Because ALiBi is natively an additive bias-based method, it requires an exponential operation to derive the actual positional coefficient; thus, we adapted the PCM-V method into \textbf{PCM-V-Exp} for compatibility. We observed that while ALiBi-B-L achieves competitive baseline performance, the incremental gain from PCM-V-Exp is marginal.

\vspace{.5em}

In a follow-up experiment, we refactored ALiBi-B-L into a multiplicative framework (\textbf{ALiBi-B-L-Mul}) and applied the standard \textbf{PCM-V} optimization. This configuration yielded a significant increase in accuracy, matching the performance of our \textbf{AGF + PCM-V} approach (see Table \ref{tab:alibi_comp}).

\begin{table}[htbp]
\centering
\caption{Performance Comparison of Positional Mechanisms and PCM-V Optimizations.}
\label{tab:alibi_comp}
\begin{tabular}{lc}
\hline
\textbf{Model Configuration} & \textbf{Validation Accuracy} \\ \hline
ALiBi-B-L                    & 70.4849 \\
ALiBi-B-L + PCM-V-Exp        & 70.5174 \\
ALiBi-B-L-Mul + PCM-V        & \textbf{70.7582} \\ \hline
\end{tabular}
\end{table}

\vspace{.5em}
These results provide compelling evidence that \textbf{dual multiplication} represents the superior architectural choice for relative positional encoding. Additive or bias-based approaches act as a form of \textbf{``premature integration"}, introducing data distortions that preclude the benefits of post-normalization corrections like PCM-V. Notably, while this optimization is seamlessly compatible with AGF and relative schemes, it remains inaccessible to Vanilla Transformers and other models confined to absolute positional encodings.

\subsection{Other Tricks}
The modular nature of the AGF framework allows for the integration of auxiliary optimization techniques. We explore two primary extensions:

\begin{itemize}
    \item \textbf{Hybrid Positional Representation:} By concurrently employing AGF and Absolute Positional Encoding (PE), the model can potentially decouple positional tasks across the $LC_1$, $LC_2$, and $LC_3$ hierarchies. In this schema, each component focuses on fitting specific functional granularities. While PE offers minimal computational overhead, its tendency to entangle positional and semantic features may introduce representational interference.
    \item \textbf{Score Calculation Optimization (SCO):} We investigated an alternative normalization for Attention scores, substituting the standard $\sqrt{d}$ scaling with a dynamic norm:
    \begin{equation}
        \text{Score}(q_m, k_n) = \frac{\boldsymbol{q}_{m}^{\top} \boldsymbol{k}_{n}}{\|\boldsymbol{k}_{n}\|}
    \end{equation}
    Empirical observations suggest that while SCO provides benefits in specific configurations, its generalizability requires further investigation.
\end{itemize}

As shown in Table \ref{tab:hybrid_results}, certain combinations achieved state-of-the-art accuracy within our experimental setup.

\begin{table}[htbp]
\centering
\caption{Validation Accuracy Across Various Optimization Combinations.}
\label{tab:hybrid_results}
\begin{tabular}{lc}
\hline
\textbf{Configuration} & \textbf{Validation Accuracy} \\ \hline
Vanilla Transformer (QKV)             & 70.5911 \\
AGF-M + SCO + PE                        & 70.8049 \\
AGF-M + SCO + PCM-V + PE     & \textbf{70.9213} \\ \hline
\end{tabular}
\end{table}

\vspace{.5em}

The emphasis of this study is the formal validation of the AGF architecture; these results serve to highlight the broad spectrum of architectural refinements made possible by the decoupling of positional information.
\section{Why AGF Works!}
\subsection{What is Attention?}

In natural language, Part-of-Speech (PoS) analysis---encompassing categories such as nouns, verbs, adjectives, and adverbs---reveals intricate hierarchical sub-classifications and syntactic frameworks (e.g., active vs. passive voice). Distinct PoS tags and syntactic structures exhibit stable dependency and modification patterns: adjectives typically modify nouns, while adverbs of manner are constrained to modify verbal phrases. Furthermore, the relative positioning of nouns and verbs serves as a primary indicator of thematic roles, such as subject and object. Fundamentally, these linguistic "rules" are equivalent to the \textbf{Attention mechanism}, which captures the statistical regularities of token pairs that frequently exhibit co-occurrence, linkage, or modification relationships.

\begin{figure}[h]
    \centering
    \includegraphics[width=0.9\linewidth]{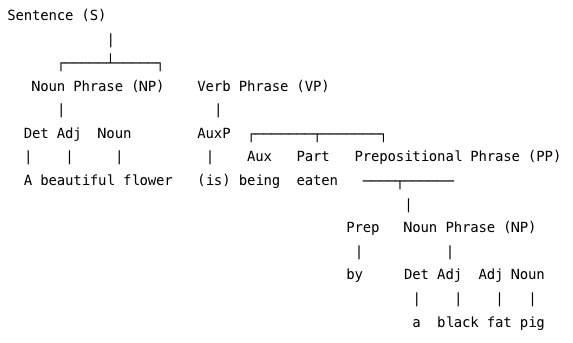}
    \caption{Part-of-Speech  VS Attention}
    \label{fig:placeholder}
\end{figure}

\vspace{.5em}

Crucially, every established linguistic rule can be refined into finer granularities. For instance, the category of "nouns" can be subdivided into semantic clusters such as \textbf{flora and fauna}, inanimate objects, or abstract concepts. In a sequence containing the adjective \textit{"beautiful"}, the transition probability to the noun \textit{`girl'} is empirically much higher than to other nouns like \textit{`pig'} or \textit{`boy'}. To formalize this, we define \textit{`beautiful'} as the Query ($Q$) and the semantic class of \textbf{flora and fauna} as the Value ($V$) to investigate the governing principles of the Attention mechanism in this context.

\vspace{.5em}

Given a massive corpus, the probability distribution of words following \textit{`beautiful'} can be modeled as a probability density. We define the Attention score associated with a specific word at a relative distance of one as $P(\text{pos}=1)$, where the integral of the distribution is normalized to unity:
\begin{equation}
    \sum P(\text{pos}=i) = 1
\end{equation}

\vspace{.5em}

\begin{figure}[h]
    \centering
    \includegraphics[width=.9\linewidth]{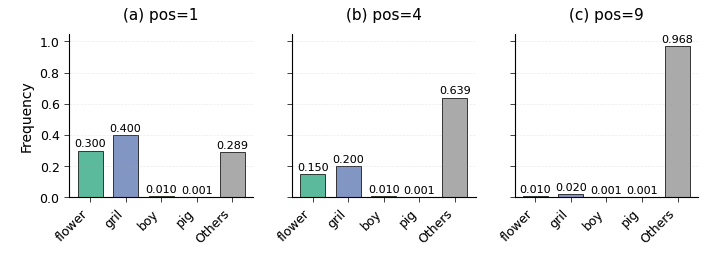}
    \label{fig:placeholder}
     \caption{Frequency Distribution of Words Following `beautiful'
}
\end{figure}
Consider the scenario where the relative distance increases to, for example, $pos=9$:
\begin{center}
    \texttt{"(beautiful) x x x x x x x x (?)"}
\end{center}
This might occur in a sentence such as: \textit{``A beautiful flower is being eaten by a black fat pig."} In this instance, the probability density $P(\text{pos}=9)$ for the token \textit{`pig'} may exceed that of \textit{`girl'}. This phenomenon stems from the fact that the syntactic dependencies underlying Attention are governed by specific structural constraints. Once the distance exceeds these boundaries, the probability of encountering stochastic, unrelated tokens increases, causing the specific Attention score to attenuate as the total probability remains normalized to unity.
 
\vspace{.5em}
However, if the modification relationship remains intact despite increased distance, the intervening tokens serve as multi-scale refinements of the core concepts. We illustrate this with the following examples:
\begin{itemize}
    \item \textit{``A \textbf{(beautiful)} and elegant \textbf{(girl)} is singing."}
    \item \textit{``A \textbf{(beautiful)}, but tired \textbf{(girl)} is working."}
\end{itemize}
In these cases, while supplementary descriptors increase the relative displacement, the fundamental syntactic dependency persists.
\vspace{.5em}

By performing statistical analysis across a massive corpus, we can derive the \textbf{PASL (Probability of Attention’s Sequence Length)}, denoted as $P_{ASL}(i)$ for $i \geq 0$, representing the probability density of a syntactic structure with length $2+i$. The expectation of a consistent decline is rooted in the principles of \textbf{Shannon Entropy} and \textbf{Huffman Coding}. In Information Theory, to achieve optimal compression, high-frequency symbols are assigned shorter bitstrings. The theoretical limit, defined by the entropy $H(X)$, is:
\begin{equation}
    H(X) = -\sum_{i=1}^{n} p(x_i) \log_2 p(x_i)
\end{equation}

\vspace{.5em}
Human language adheres to this principle of \textbf{linguistic economy}. Humans instinctively utilize the briefest possible expressions for the most frequent events, adding complexity only for low-frequency details. Should a longer construct $P_{ASL}(i+1)$ become more frequent than its shorter counterpart $P_{ASL}(i)$, linguistic evolution (often through abbreviations or lexicalization) would eventually favor a more concise representation. Consequently, the global trend of PASL is inevitably a declining curve.

\vspace{.5em}

However, there are numerous choices for the declining curve, such as exponential curves or power-law curves. Which one is truly the most appropriate?
\vspace{.5em}

\begin{figure}[H]
    \centering
    \includegraphics[width=.7\linewidth]{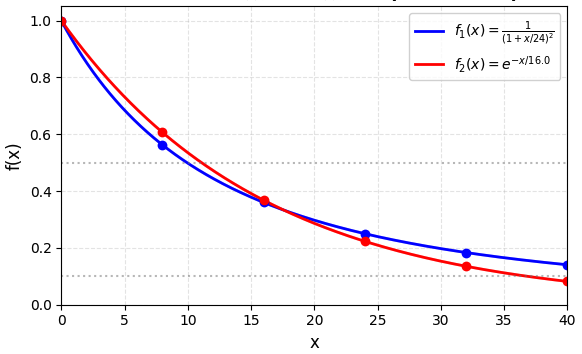}
    \caption{Power vs Exp}
    \label{fig:placeholder}
\end{figure}

\subsection{Power-Law}
In the domain of reliability engineering, the \textbf{Duane Model}, introduced in 1962 by J.T. Duane, serves as the inaugural widely-adopted reliability growth framework. Its fundamental premise is that, given continuous improvement, the cumulative failure rate and cumulative testing time maintain a linear relationship in a log-log coordinate system. This is categorized as a continuous model, assuming smooth, incremental growth without discrete structural mutations. The relationship is formulated as:
\begin{equation}
    \ln(\text{MTBF}) = -\ln a + m \ln t
\end{equation}

The probability density of varying sequence lengths in our Attention mechanism can be conceptualized as a reliability growth problem centered on "requirement satisfaction." A sentence of length $\text{SeqLen} = i+1$ is triggered only when the preceding length $\text{SeqLen} \le i$ proves insufficient for the required descriptive precision. By equating probability density to "failure occurrence," the system stabilizes as descriptive needs are progressively met. Consequently, the \textbf{PASL} problem is a manifestation of reliability growth, which follows a \textbf{power-law} distribution.

\vspace{.5em}
Alternatively, the \textbf{Learning Curve} provides another interpretative lens. The Cumulative Distribution Function (CDF) of the PASL can be viewed as the system's "perfection score." As the PASL length increases, its coverage of linguistic events expands toward a 100\% asymptote. This cumulative form aligns with classic learning curves, which are empirically governed by power laws. To validate this, we perform a regression analysis on the scores obtained from 10 training epochs in our study:
\newline
\fontsize{8}{10}$[62.9851,67.6618,69.025,69.7529,70.1426,70.4782,70.6026,70.7271,70.7603,70.9213]$
\newline
\vspace{.5em}
\begin{figure}[h]
    \includegraphics[width=0.8\linewidth]{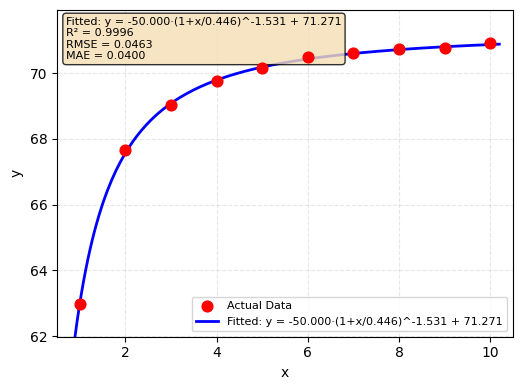}
    \centering
    \caption{Learning Curve}
    \label{fig:placeholder}
\end{figure}
\newline
Remarkably, the fitting analysis even yields the theoretical performance ceiling for our model, which is determined to be \textbf{71.271}.

\vspace{.5em}
The \textbf{Power Law} is a fundamental principle observed across various domains, from the \textbf{Matthew Effect} to the \textbf{Pareto Principle}. Notably, \textbf{Zipf (1932)} demonstrated that word frequencies in natural language adhere to a power-law distribution. We contend that the \textbf{PASL} problem is no exception and is fundamentally governed by power-law dynamics.

\vspace{.5em}
However, a critical question arises: if PASL is a power-law problem, why do exponential-based fittings still perform effectively? This observation is mirrored in the \textbf{KERPLE} study, where \textbf{KERPLE-log} (Power-Law) only marginally outperformed \textbf{KERPLE-power} (Exponential). 

\vspace{.5em}
The explanation lies in the geometric proximity of these curves at localized scales. At relatively short distances, exponential and power-law decays exhibit near-identical trajectories. Their divergence becomes pronounced only in the "long-tail" region, where the power law attenuates gradually while the exponential curve drops precipitously. In the context of \textbf{Attention score} computation, the mechanism prioritizes high-magnitude components (the "main energy"). Redundant, infinitesimal tail values are often negligible or even detrimental. In fact, such values must be suppressed via \textbf{Softmax} to prevent the cumulative noise of distant tokens from interfering with the primary semantic signals.

\vspace{.5em}

Consequently, we hypothesize that while the Power Law provides a more faithful representation of the \textbf{PASL} phenomenon, the exponential function serves as a viable approximation. Its inherent suppression of distal "tail" data may, paradoxically, facilitate more robust \textbf{Softmax} normalization in Attention mechanisms.

\subsection{Gravitational Field}

The ubiquity of \textbf{Power Laws} across diverse disciplines is attributed to various underlying phenomena, each characterized by specific theoretical frameworks such as \textbf{preferential attachment} in complex network analysis, \textbf{self-organized criticality}, and the emergence of \textbf{scale-free networks}. 

\vspace{.5em}

In the following section, we deviate from a strictly formalistic approach to provide a more \textbf{intuitive heuristic}. This descriptive derivation aims to elucidate the underlying logic of the \textbf{PASL} problem and the rationale behind its power-law nature.

\subsubsection{The Expanding Sphere Model}
The \textbf{PASL} distribution, $P_{ASL}(pos=r)$, exhibits a monotonic decay. However, a critical observation is that the incremental growth at distance $r$ is not exclusively derived from the immediate precursor $P_{ASL}(r-1)$. 
Specifically, while a phrase of length $SeqLen=r$ is highly probable to be a \textbf{recursive refinement} or augmentation of a $SeqLen=r-1$ structure, it may also originate from earlier states such as $r-2$ or $r-3$. Theoretically, the transition probability diminishes as the interval increases. As the \textbf{scanning radius} $r$ expands, these multi-stage increments superimpose, resulting in a complex network of internal dependencies and structural correlations.

\begin{figure}[h]
    \centering
    \includegraphics[width=1\linewidth]{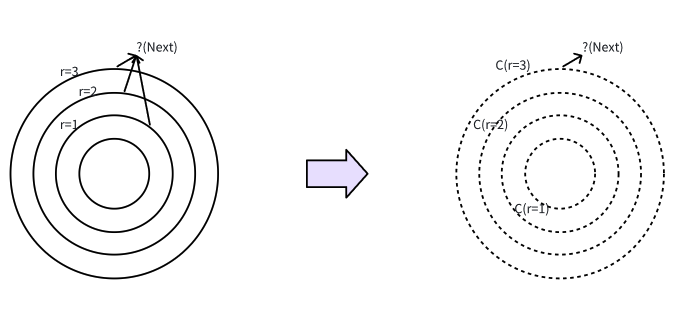}
    \caption{Expanding Sphere}
    \label{fig:placeholder}
\end{figure}
\vspace{.5em}
To formalize this recursive dependency, we define a virtual construct $C\{R\}$ with a scanning radius $R > 0$. The state $C\{R=r\}$ represents the integration of all cumulative increments up to the boundary $r$. We introduce an increment function $G(r, x)$, where $x > 0$, to denote the gain associated with an additional length $x$ given the state $C\{R=r\}$. 

\vspace{.5em}

To maintain structural continuity during the expansion from $C\{R=r\}$ to $C\{R=r+1\}$, the system must satisfy the consistency constraint:
\begin{equation}
    P_{asl}(r + x) = G(r, x) = G(r+1, x-1)
\end{equation}
By defining a basis function $F(x)$, we obtain the relation $G(r, x) = F(r + x)$. This conceptualizes the linguistic structure as an expanding sphere bound by an \textbf{elastic membrane} whose properties scale with the radius. 

\vspace{.5em}

We further define a reciprocal ``surface area" function $S(x) = 1/F(x)$. Our objective is to maximize the informational coverage for each unit increase in $R$. Since the incremental density is inversely proportional to $S(x)$, the principle of efficiency dictates the minimization of $S$ at any given radius. According to the \textbf{isoperimetric inequality} in multi-dimensional space, a sphere minimizes surface area for a fixed volume. This underlying optimization mirrors the algebraic inequality $(x+1)(x-1) < x^2$, suggesting that $C\{R\}$ functions as a spherical field where $S$ represents its surface area. This leads directly to the power-law formulation:
\begin{equation}
    G(r, x) = \frac{1}{(r+x)^k}, \quad k > 0
\end{equation}

\vspace{.5em}

This framework establishes a profound analogy with physical phenomena, such as the \textbf{inverse-square law} of luminous flux or \textbf{Newtonian gravitation}. Consequently, designating the smooth decay of Attention intensity as the \textbf{AGF (Attention's Gravitational Field)} provides both a precise and intuitive physical grounding. AGF serves as the fundamental mechanism underlying the power-law emergence in the PASL problem.

\subsubsection{Deep Smoothing and Hierarchical Evolution}

From an alternative perspective, let $S(i)$ represent successive increments in the system. An \textbf{exponential distribution} assumes a constant decay ratio $k$, such that $S(i+1) = S(i) \cdot k$. 

\begin{figure}[h]
    \centering
    \includegraphics[width=0.7\linewidth]{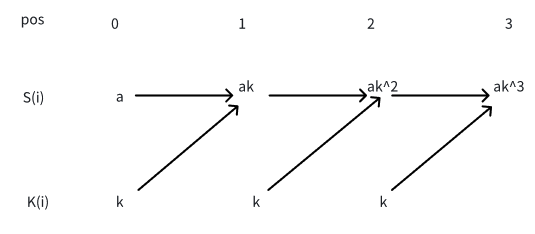}
    \caption{exponential distribution}
    \label{fig:placeholder}
\end{figure}
However, in high-dimensional complex environments, physical variables typically exhibit deep-seated hierarchical dependencies. As the magnitude $S(i)$ attenuates, it is empirically more plausible that the decay coefficient $k$ itself should be subject to attenuation. This recursive dependency implies that surface-level features sustain higher sensitivity to change, while deeper structural layers exhibit increased stability.

The distinction is formalized by examining the derivative properties. For the \textbf{exponential family}, the rate of change is static and self-similar:
\begin{align}
    f(x) &= e^x \\
    f'(x) &= e^x \\
    f''(x) &= e^x
\end{align}
This "one-size-fits-all" decay lacks the capacity to model multi-scale processes. In contrast, \textbf{power-law functions} exhibit a systematic, layered evolution from the superficial to the profound:
\begin{align}
    f(x) &= x^k \\
    f'(x) &= k \cdot x^{k-1} \\
    f''(x) &= k(k-1) \cdot x^{k-2}
\end{align}

This mathematical progression captures what we term \textbf{Deep Smoothing}. Compared to the idealized attenuation of exponential functions, the following power-law formulation for the dynamic decay ratio $K(i)$ provides a more faithful representation of hierarchical smoothing:
\begin{equation}
    K(i) = \frac{S(i+1)}{S(i)} = \left( \frac{d+i}{d+i+1} \right)^k, \quad i > 0, d > 0
\end{equation}

\section{Conclusion}

The \textbf{PASL} phenomenon in Attention mechanisms is fundamentally driven by the duality of human cognitive economy and ingenuity. Efficiency dictates the formulation of concise objectives, while intelligence enables their realization through structured complexity. Within this framework, the \textbf{power law} serves as the primary mathematical vehicle for goal attainment. This principle extends to reliability engineering, industrial yield optimization, and the training of artificial intelligence. We designate the trajectory of increasing system capability as the \textbf{IGC (Intelligence Growth Curve)}, and the corresponding distribution of resolved complexities as the \textbf{P-IGC (Pain of Intelligence-Growth Curve)}. Both functions are inherently characterized by power-law dynamics.
\vspace{1em}

\begin{figure}[h]
    \centering
    \includegraphics[width=.7\linewidth]{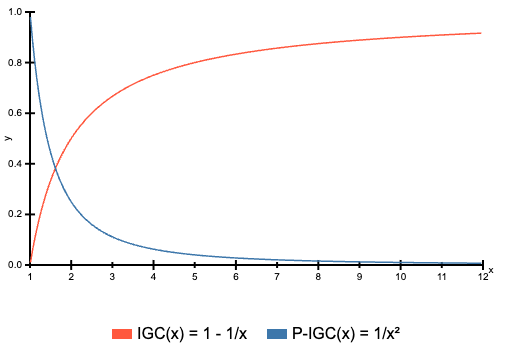}
    \caption{IGC and P-IGC}
    \label{fig:placeholder}
\end{figure}

The ubiquity of the power law in modeling ultra-complex systems---ranging from celestial mechanics to stochastic linguistic events---stems from its unique mathematical properties. It effectively captures multi-variable correlations through multiplicative interactions while representing optimal configurations under structural constraints, much like the geometric efficiency of a sphere. Moreover, the systematic evolution of its higher-order derivatives aligns with the natural progression from superficial observations to deep-seated structural regularities.

\vspace{.5em}

In conclusion, this study has systematically analyzed the intersection of natural language properties and Attention mechanisms. By synthesizing these perspectives, we have derived a theoretical framework that we consider the most robust explanation for distance-based decay in LLMs. The insights provided herein offer a novel foundation for future endeavors in model optimization and the pursuit of AI interpretability.

\section*{Acknowledgements}

The author would like to acknowledge the invaluable assistance provided by AI tools, including \textbf{Yuanbao-AI} and \textbf{Gemini}, which greatly facilitated the research process. Their capabilities are truly remarkable.

\vspace{0.5em}

Special thanks are also due to the authors of the \textbf{KERPLE} \cite{chi2022kerple}paper for their rigorous and in-depth evaluations. By providing comprehensive comparative analyses across extensive datasets and various fitting functions, their exceptional work saved us a tremendous amount of effort and time.

\printbibliography[resetnumbers=True]

\end{document}